\documentclass[submission,copyright,creativecommons]{eptcs}
\providecommand{\event}{41st International Conference on Logic Programming (ICLP 2025)} 
\usepackage{listings}
\usepackage{tikz}
\usetikzlibrary{shapes.geometric, arrows, positioning}
\usepackage{etoolbox} 
\hypersetup{hidelinks}
\usepackage{multicol}
\AtBeginEnvironment{align}{\small}
\AtBeginEnvironment{gather}{\small}
\usepackage{caption}
\usepackage{listings}

\usepackage{fancyvrb}
\usepackage{multirow}
\usepackage{array}
\tikzset{
  process/.style={
    draw,
    fill=white!20,
    text centered,
    minimum height=2em
  }
}
\usepackage{amsmath}
\usepackage{multicol}
\AtBeginEnvironment{lstlisting}{\scriptsize}
\usepackage{iftex}
\usepackage{amsmath}
\ifpdf
  \usepackage{underscore}         
  \usepackage[T1]{fontenc}        
\else
  \usepackage{breakurl}           
\fi

\title{Hybrid MKNF for Aeronautics Applications: Usage and
Heuristics}
\author{
  Arun Raveendran Nair Sheela
    \institute{Université Clermont Auvergne, LIMOS}
    \institute{Thales}
  \and
  Florence De Grancey
    \institute{Thales}
  \and
  Christophe Rey
    \institute{Université Clermont Auvergne, LIMOS}
    \institute{CNRS, France}
  \and
  Victor Charpenay
    \institute{École des Mines de Saint-Étienne, LIMOS}
    \institute{CNRS, France}
}

\def\titlerunning{Hybrid-MKNF for Aeronautics Applications: Usage and
Heuristics}
\def\authorrunning{A.R.N. Sheela, F. De Grancey, C. Rey \& V. Charpenay}
\begin{document}
\maketitle

\begin{abstract}
The deployment of knowledge representation and reasoning technologies in aeronautics applications presents two main challenges: achieving sufficient expressivity to capture complex domain knowledge, and executing reasoning tasks efficiently while minimizing memory usage and computational overhead. An effective strategy for attaining necessary expressivity involves integrating two fundamental KR concepts: rules and ontologies. This study adopts the well-established KR language Hybrid MKNF owing to its seamless integration of rules and ontologies through its semantics and query answering capabilities. We evaluated Hybrid MKNF to assess its suitability in the aeronautics domain through a concrete case study. We identified additional  expressivity features  that are crucial for developing aeronautics applications and proposed a set of heuristics to support their integration into Hybrid MKNF framework.
\end{abstract}

\section{Introduction}

The aeronautics industry is acknowledged as a safety-critical domain where system failures can have serious safety implications for stakeholders. Therefore, any software system integrated into this sector must exhibit trustworthy attributes such as reliability, robustness, and explainability to prevent such consequences. Moreover, many software systems in this field operate on resource-constrained hardware  such as microcontrollers. In this context, Knowledge Representation and Reasoning (KRR) shows significant promise as it enables systems to make transparent and explainable decisions. With the assistance of domain experts, knowledge specific to aeronautics applications has been captured using formal representation languages, such as ontologies. This knowledge encompasses critical operational information and can be combined with real-time data to facilitate timely decision-making through logical reasoning. However, adapting KRR to the aeronautics domain presents several significant challenges. These include ensuring sufficient expressivity to encapsulate all complex domain knowledge and effectively executing the necessary reasoning tasks with limited computational resources and minimal latency.

 An effective approach to addressing the expressivity challenge involves integrating rules and ontologies, which is a well-researched area. Ontologies, typically based on Description Logics (DLs), operate under the open-world assumption (OWA), meaning that the absence of information does not equate to its negation. They encapsulate general domain knowledge through concepts and their relationships \cite{Baader2003TheDL}. In contrast, Logic Programming (LP) rules function under the closed-world assumption (CWA), assuming that what is not known to be true is considered false. These rules represent knowledge through if-then relationships \cite{Schulze2016HandbookOL}. Owing to their complementary features, combining DLs and LP rules enhances the expressiveness of the resulting KRR language, which is known as a hybrid knowledge base. Various methodologies have been proposed for creating hybrid knowledge bases, each offering unique advantages and trade-offs \cite{Knorr2021OnCOsurvey}, \cite{Drabent2010HybridRWformalsurvey}, \cite{ruleontologyrecsurvey},  \cite{featurecombingruleontolgoyt}. In this study, we explore Hybrid MKNF language, which provides a unified logic that integrates the semantics of DLs and LP \cite{hybridMknfintroduction}. Both well-founded semantics \cite{Knorr2008ACW} and answer-set semantics \cite{mknf+} have been proposed for  Hybrid MKNF. Given that complex query answering constitutes the primary reasoning task within the aeronautics domain and that well-founded semantics offers greater computational efficiency for handling it, we adopted this semantics in our approach.

The primary contribution of this study lies in exploring Hybrid MKNF under well-founded semantics and its implementation in the aeronautics sector, along with its practical adaptability through a comprehensive evaluation. To assess this, we conducted a series of experiments focusing on preprocessing delays, query answering performance, and resource utilization. We also acknowledge that Hybrid MKNF under well-founded semantics does not inherently support all expressive features required for modeling complex aviation scenarios. To address this limitation, we define a set of core expressive needs using a semi-realistic use case, and propose heuristics to demonstrate how these features can be effectively integrated into Hybrid MKNF framework. In addition, we address future research directions to further enhance the features of Hybrid MKNF in an aeronautics context.

Section 2 presents a semi-realistic use case within the aeronautics sector to illustrate the concepts discussed in this study. Section 3 provides an overview of Hybrid MKNF framework, introduces the NoHr reasoner, and explains its efficient usage in this context.  Section 4 proposes heuristics to enhance expressiveness with classical negation and integrity constraints. Section 5 describes the implementation of these features. Section 6 reviews related work and justifies the choice of Hybrid MKNF. Finally, Section 7 concludes the study and outlines future directions.

\section{Use Case}
To address a use case in the aeronautics domain requiring the integration of DLs and LP rules, we propose the NOTAM-Aware Reasoning System (NARS).
Before departure, pilots are tasked with manually reviewing all Notice to Airmen (NOTAMs) messages, which are official notifications issued to communicate time-sensitive changes in airspace status, environmental conditions, runway closures, and other relevant factors. This review aims to ensure that all NOTAMs are consistent with flight plans. Below is an example of a preprocessed NOTAM and flight plan:

\hspace{-1em}\begin{minipage}{0.4\textwidth}
  
\begin{lstlisting}
Notam001 :hasaffectedBy rw1;  
         :hasRunwayStatus :Closed;
         :hasAirport LFPG;
         :haStartTime "2025-04-18T06:00:00Z";
         :hasValidEndTime "2025-04-18T12:00:00Z";
         :hasIssuedBy :ATC_Agency1 .
\end{lstlisting}
 
    \end{minipage}
      \begin{minipage}{0.5\textwidth}
  
\begin{lstlisting}
Flight:AF123 :hasDepartureAirport LFBO ;
             :hasDestinationAirport LFPG ;
             :hasDepartureTime "2025-04-18T06:00:00Z";
             :hasArrivalTime "2025-04-18T07:30:00Z";
             :landingRunway rw1 ;
             :hasAlternateAirport ex:LFOB . 
\end{lstlisting}
 
    \end{minipage}
    
For example, if a NOTAM indicates that the designated runway is unavailable at the flight's scheduled arrival time, the pilot must either request a revised flight plan or delay the flight. Manually analyzing each NOTAM to extract such information and making appropriate decisions, whether by the pilot or supporting staff, is both time-consuming and prone to error. To address this challenge, we propose NARS that assists pilots by automatically providing recommendations based on relevant NOTAMs. Some of the key operations performed by NARS include the following:
\begin{enumerate}
    \item Decisions should be guided solely by valid NOTAMs. Any NOTAM that is invalid or contradictory, due to missing critical data or conflicting information from different authorities, must be disregarded, and pilots should be notified of its invalidity.

    \item  If the current alternate airports are found to be unavailable due to constraints reported in NOTAMs, it is necessary to recommend the designation of new alternate airports.

    \item In emergency landing situations, nearby airports are suggested along with runways that meet the necessary parameters for safely landing the aircraft. One crucial parameter is the runway length, which must exceed the aircraft’s required landing distance.
    
\end{enumerate}

The knowledge base for NARS can be constructed using Hybrid MKNF. On the DL side, static and hierarchical domain knowledge, such as airport infrastructure, different types of runways, and airspace zones can be represented. This structured knowledge provides a formal vocabulary to describe entities and their relationships. For example, stating that "\textit{a runway is part of an airport}" or "\textit{runway has an instrument landing system (ILS)}". In contrast, LP handles dynamic knowledge, which is encoded as conditional if-then rules. For instance, a rule might specify that "\textit{if a runway is closed during the scheduled landing time, then an alternate runway should be proposed}" or "\textit{if critical equipment at an airport is out of service, the flight operations team must be notified}". Specific instances, such as individual NOTAMs or flight plan, can be formalized as assertions.
\section{Hybrid MKNF}
Hybrid MKNF, as introduced in \cite{hybridMknfintroduction}, is based on the logic of Minimal Knowledge and Negation-as-a-Failure (MKNF), an extension of first-order logic that incorporates two modal operators, namely \textbf{K} and \textbf{not} originally described in \cite{MKNF1}. The \textbf{K} operator denotes knowledge known by the system, whereas the \textbf{not} operator signifies knowledge not known by the system, functioning analogously to default negation in LP and facilitating closed-world reasoning. A Hybrid MKNF under well-founded semantics is defined as \(KB= (O,P)\) where O is a DL ontology and P is the set of  MKNF rules of the form \cite{Knorr2011LocalCW}:
\vspace{-.3em}
\begin{equation}
   \mathbf{K} \ H  \leftarrow \mathbf{K} \  A_1, \dots,\mathbf{K} \  A_m, \mathbf{not} \ B_1, \dots, \mathbf{not} \ B_n.
\end{equation}
where \(H\), \(A_i\) for \(1 \leq i \leq m\),  and \(B_j\) for \(1 \leq j\leq n\) are first-order atoms. The rule (1) can be read as "\textit{if all \(A_i\) are known to hold and all \(B_j\) are not known to hold then \(H\) is known to hold}". We consider a simplified scenario from the use case, in which airport and runway availability is inferred from NOTAM data, and operational flight recommendations are derived using a Hybrid MKNF knowledge base,  \(KB= (O,P)\):
 \vspace{-2.5em}
\begin{multicols}{2}
\begin{align}
\text{Airport} \sqcap \forall \text{hasRwy.CldRwy} &\sqsubseteq \text{CldAirport} \label{eq:closed-airport} \\
\text{Airport} \sqcap \exists \text{hasRwy.OpnRwy} &\sqsubseteq \text{OpnAirport} \label{eq:open-airport}
\end{align}

\columnbreak

\begin{align}
\text{OpnRwy} &\sqsubseteq \text{Rwy} \label{eq:open-runway} \\
\text{CldRwy} \sqcap \text{OpnRwy} &\sqsubseteq \bot \label{eq:disjoint-runways}
\end{align}
\end{multicols}
\vspace{-2em}

   \begin{gather}
 \mathbf{K} \ \text{CldRwy}(X) \gets \mathbf{K} \  \text{NOTAM}(X), \mathbf{K} \ \text{affdBy}(X, Y), \mathbf{K} \ \text{Rwy}(Y),\ 
   \mathbf{K} \ \text{opStat}(X, \text{closed}). \\
\mathbf{K} \ \text{OpnRwy}(X) \gets \mathbf{K} \ \text{Rwy}(X),\ \mathbf{not}\ \text{CldRwy}(X). \\
\mathbf{K} \ \text{recommendDelay}(X) \gets \mathbf{K} \ \text{Flight}(X),\mathbf{K} \ \text{hasDepartureAirport}(X,Y), \mathbf{not} \ \text{OpnAirport}(Y).
\end{gather}

Rule (7) infers that \textit{rwy(rw1)} is considered an \textit{OpnRwy(rw1)} by default, unless there is explicit evidence indicating that \textit{rwy(rw1)} is a \textit{CldRwy(rw1)}. The predicate \textit{CldRwy(rw1)} can be derived when there exists a NOTAM that affects the specific runway, as described in rule (6). This can be expressed using the \textbf{not} operator: \textbf{not}\textit{ CldRwy(X)} signifies that it is not known that a runway is closed.
 This allows the inference that, in the absence of a NOTAM message indicating a 
closure, the runway remains operational. Furthermore, by asserting the following additional 
facts: $\textit{hasRwy(lfbo, rw1)}$ and $\textit{Airport(lfbo)}$
and applying rule (7) in conjunction with DL axiom (3), it can be concluded that 
$\textit{Airport(lfbo)}$ qualifies as an $\textit{OpnAirport}$.

The semantics of Hybrid MKNF can be formalized  by transforming the knowledge base into a first-order formula with modal operators \cite{Knorr2011LocalCW}. Consider  \(KB_1= (O_1,P_1)\) where \(O_1\) contain DL axiom (4) and an assertion \textit{Rwy(rw1)}, and \(P_1\) contain a rule (7). The first-order translation of \(KB_1\), denoted as \(\pi(KB_1)\) is given by \( \mathbf{K} \pi(O_1) \wedge  \pi(P_1)\):
\vspace{-.6em}
\begin{gather}
\forall X (  \mathbf{K} \ \text{Rwy}(X) \gets  \mathbf{K} \ \text{OpnRwy}(X)). \\ 
\forall X( \mathbf{K} \ \text{OpnRwy}(X) \gets  \mathbf{K} \ \text{Rwy}(X),\  \mathbf{not}\ \text{CldRwy}(X)).
\end{gather}
A model of an MKNF formula is defined as a maximal set of interpretations that satisfy
the formula, such that no proper superset of this 
set also satisfies the formula \cite{MKNF1}. \(KB_1\) is satisfiable if there exists a MKNF model for \(\pi(KB_1)\) and  \(KB_1\)  entails a first-order formula \(\psi\), denoted as  
\(KB_1\models \psi\) if and only if \(\pi(KB_1) \models_{MKNF} \psi\). 
The semantics of \(\pi(KB_1)\) can be explained using either answer set \cite{mknf+} or well-founded semantics \cite{Knorr2011LocalCW}.
In the field of aeronautics, well-founded semantics is essential because they establish a single three-valued model comprising \emph{true}, \emph{false}, and \emph{undefined} atoms, thereby facilitating an efficient top-down query answering.
 To explain well-founded semantics for Hybrid MKNF, a three-valued MKNF interpretation is defined as a pair \((M, N)\), where both \(M\) and \(N\) are sets of first-order interpretations with \(N \subseteq M\). To avoid unintended consequences when integrating DL with LP, the standard name assumption (SNA) is enforced, ensuring each constant uniquely identifies an individual and equality is explicitly handled as a congruence relation \cite{mknf+}.

Truth evaluation of first-order atoms with modal operators in relation to the three-valued MKNF model, proceeds as follows.
An atomic formula \(P(t_1, \dots, t_n)\) is evaluated as \emph{true} under a first-order interpretation \(I\) if the tuple \((t_1^I, \dots, t_n^I)\) is an element of the interpretation of the predicate \(P\) in \(I\), denoted \(P^I\); formally, if \((t_1^I, \dots, t_n^I) \in P^I\). Otherwise, the formula is evaluated as \emph{false} when \((t_1^I, \dots, t_n^I) \notin P^I\).  The implication \(\varphi_1 \supset \varphi_2\) is evaluated as \emph{true} if the truth value of \(\varphi_2\) is greater than or equal to that of \(\varphi_1\) with respect to the order \(f < u < t\), and \emph{false} otherwise.  The truth value of \(\mathbf{K} \ \varphi\) is \emph{true} if \(\varphi\) evaluates to true in all interpretations in \(M\),  \emph{false} if \(\varphi\) evaluates to false in some interpretation in \(N \subseteq M\), and \emph{undefined} otherwise. Similarly, \(\textbf{not } \varphi\) is \emph{true} if \(\varphi\) is false in some interpretation in \(N \subseteq M\), \emph{false} if \(\varphi\) is true in all interpretations in \(M\), and \emph{undefined} otherwise \cite{Knorr2011LocalCW}.

An MKNF interpretation pair \((M, N)\) satisfies a closed MKNF formula \(\varphi\), denoted \((M, N) \models \varphi\), if and only if
\(
(M,N)(\varphi) = t,
\)
meaning that all rules within \(\varphi\) evaluate to true under this interpretation. An MKNF interpretation pair \((M, N)\) is considered a three-valued MKNF model for a formula \(\varphi\) if it satisfies two key conditions. First, it must satisfy the formula \(\varphi\) itself. Second, it must be minimal in the sense that one cannot extend \(M\) or \(N\) to larger interpretation pairs that still satisfy \(\varphi\).
 The only 
three-valued  model of  \(KB_1\): \textit{M =  \{\{Rwy(rw1), OpnRwy(rw1)\}, \{Rwy(rw1), OpnRwy(rw1), 
 CldRwy(rw1)\}\}, \ N =  \{\{Rwy(rw1), OpnRwy(rw1)\}, \{Rwy(rw1), OpnRwy(rw1), 
 CldRwy(rw1)\}\}}. 

\paragraph{Query Answering:}
The alternating fixed-point method provides a bottom-up strategy for calculating the three-valued, well-founded model of a Hybrid MKNF knowledge base \cite{Knorr2011LocalCW}. Initially, this model must be computed using the proposed bottom-up approach. Once computed, it can be stored in a database for querying purposes.  However, this method is impractical for large knowledge bases, particularly when only specific parts of information are relevant to answering a query. To address this issue, a query answering algorithm called SLG(O) resolution has been proposed, which is both sound and complete with respect to the well-founded Hybrid MKNF model. SLG(O) resolution is a goal-directed query answering that employs SLG resolution to manage LP rules, where O acts as the oracle responsible for handling queries related to the DL component \cite{Alferes2010QueryDrivenPF}. The resolution process begins with a query that is matched against the rule head, subsequently generating new subgoals derived from the body of that rule. When a subgoal corresponds to a DL atom, the system invokes the DL oracle to determine whether the atom is entailed by DL. Throughout the process, tabling is used to store the intermediate results, thereby preventing infinite loops and ensuring termination. Before implementing SLG(O) resolution, a knowledge base doubling step must be conducted to align the algorithm with the bottom-up computation of the well-founded MKNF model \cite{Alferes2010QueryDrivenPF}.

Consider  \( KB = (O, P) \) to be a Hybrid MKNF knowledge base and in \cite{Alferes2010QueryDrivenPF} introduce new predicates \( L^d \) and \( NL \) for each predicate \( L \) appearing in \( KB \), and then define:
\begin{enumerate}
    \item \( O^d \) by substituting each concepts or role \( L \) in \( O \) by \( L^d \); and
    \item \( P^d \) by transforming each rule of form (1)
    occurring in \( P \) into two rules:

\begin{enumerate}
    \item[i] \( \mathbf{K} \ H \leftarrow \mathbf{K} \ A_1, \dots, \mathbf{K} \ A_n, \mathbf{not} \ B^{d}_1, \dots, \mathbf{not} \ B^{d}_m \) and either,

    \item [ii(a)] \( \mathbf{K} \ H^d \leftarrow \mathbf{K} \ A^{d}_1, \dots, \mathbf{K} \ A^{d}_n, \mathbf{not} \ B_1, \dots, \mathbf{not} \ B_m, \mathbf{not} \ N H \) \text{ if } \( H \text{ is a DL concept or role} \); or

        \item [ii(b)]  \( \mathbf{K} \ H^d \leftarrow \mathbf{K} \ A^{d}_1, \dots, \mathbf{K} \ A^{d}_n, \mathbf{not} \ B_1, \dots, \mathbf{not} \ B_m \) \text{ otherwise}.
\end{enumerate}
\end{enumerate}
The doubled Hybrid MKNF knowledge base is defined as $KB^d = (O, O^d, P^d)$. In this transformation, each original predicate $L$ in the knowledge base $KB$ is paired with two predicates, $L$ and $L^d$, where $L^d$ indicates the non-falsity of $L$. In addition, a new predicate $NL$, semantically equivalent to the classical negation $\neg L$, is introduced. This predicate prevents the derivation of $L^d$ when $\neg L$ is entailed in the ontology. To enforce this, $\mathbf{not} \ NL$ is added to the body of every rule, with $L^d$ as its head, ensuring that $L^d$ can only be inferred when $\neg L$ is not supported by the ontology. This transformation process is called a semi-negative transformation. It aids in verifying MKNF consistency by identifying cases where $L$ is true but $L^d$ is not. In such instances, because \(L\) is true and \(L^d\) is false, it follows that \(\neg L\) is entailed by the DL ontology O, leading to inconsistency.
\vspace{-.3em}
\subsection{NoHr Reasoner}
The \textit{NoHr} reasoner employs  SLG(O) resolution, fundamentally translating the doubled Hybrid MKNF knowledge base into an equivalent set of  rules \cite{Kasalica2020NoHRAO}. This method facilitates efficient reasoning through query answering by using \textit{XSB Prolog}. Specifically, the doubled Hybrid MKNF knowledge base, represented as \( KB = (O, O^d, P^d) \), is converted into a semantically equivalent form: \(KB = (\emptyset, P^d \cup P^O),\) where \( P^d \) is the doubled logic program, and \( P^O \) is the DL ontology translated into rules, followed by knowledge base doubling.  The NoHr reasoner employs two distinct translation methodologies to convert DL into rules. First, a classification algorithm is used to deduce the subsumption relationships, which are then translated into LP rules \cite{Ivanov2013AQTEL}. For classification, NoHr integrates three existing DL reasoners: ELK, HermiT, and Konclude, with Hermit and Konclude supporting more expressive DL constructors. Second, for the QL profile, a direct translation method was applied to convert DL axioms into Logic Programs without prior classification \cite{Costa2015NextSFQL}. We chose to focus on the classification-based translation with  the HermiT reasoner, as it supports more expressive DL constructors than QL. The various steps involved in NoHr with the classification-based translator are illustrated in figure 1.

\begin{figure}[h]
    \centering
\begin{tikzpicture}[node distance=0.7cm and 1.2cm, auto]
        \node (dl) [process] {\small DL Ontology};
        \node (loading) [process, right=of dl] {\small Loading};
        \node (normalization) [process, right=of loading] {\small Normalization};
        \node (inference) [process, right=of normalization] {\small Classification};

        \node (translation) [process, below=.6cm of inference] {\small Translation};
        \node (doubling) [process, left=of translation] {\small KB Doubling};
        \node (parsing) [process, left=of doubling] {\small Rule Parsing};
        \node (writing) [process, left=of parsing] {\small File Writing};

        \node(ruleprogram)[process, below=.6cm of translation] {\small Rule Program};
         \node (xsbload) [process, below=.6cm of writing] {\small XSB-Prolog};
        \node (xsbquery) [process, right=of xsbload] {\small Interface};
       
        \draw [->] (dl) -- node[above] {\scriptsize Step 1} (loading);
        \draw [->] (loading) -- node[above] {\scriptsize Step 2} (normalization);
        \draw [->] (normalization) -- node[above] {\scriptsize Step 3} (inference);
        \draw [->] (inference.south) -- node[left] {\scriptsize Step 4} (translation.north);
        \draw [->] (translation) -- node[above] {\scriptsize Step 5} (doubling);
        \draw [->] (doubling) -- node[above] {\scriptsize Step 6} (parsing);
         \draw [->] (doubling) -- node[below] {\scriptsize Step 10} (parsing);
        \draw [->] (parsing) -- node[above] {\scriptsize Step 7} (writing);

                        \draw [->] (parsing) -- node[below] {\scriptsize Step 11} (writing);
        \draw [->] (writing) -- node[left] {\scriptsize Step 8} (xsbload);
          \draw [->] (writing) -- node[right] {\scriptsize Step 12} (xsbload);
                        \draw [->] (ruleprogram) -- node[above] {\scriptsize Step 9} (doubling);
                 \draw [<->] (xsbquery) -- node[above] {\scriptsize Query}(xsbload);
                   \draw [<->] (xsbquery) -- node[below] {\scriptsize Response}(xsbload);

\end{tikzpicture}

     \caption{Different steps in NoHr Reasoner for EL Profile }
     \vspace{-1em}
    \label{fig:workflow}
\end{figure}
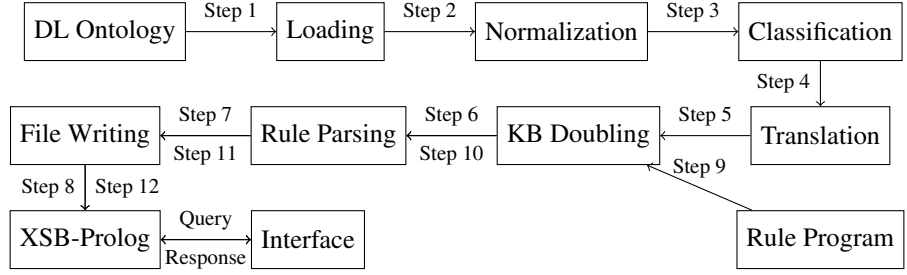
Consider DL axioms (4) and (5), which are translated into rules according to the NoHr proposition, followed by knowledge base doubling and the semi-negative transformation (Step 1 to Step 6 in figure 1, see \cite{Ivanov2013AQTEL} for more details):
\vspace{-2.5em}
\begin{multicols}{2}
\begin{align}
 \text{Rwy}(X) \gets \text{OpnRwy}(X). \\
   \text{Rwy}^d(X) \gets \text{OpnRwy}^d(X), \mathbf{not} \ \text{NRwy}(X).
\end{align}

\columnbreak

\begin{align}
\text{NCldRwy}(X) \gets \text{OpnRwy}(X). \\
  \text{NOpnRwy}(X) \gets \text{CldRwy}(X).
\end{align}
\end{multicols}
The NoHr reasoner exhibits paraconsistent  behavior through a combination of knowledge base doubling and semi-negative transformation. This approach allows reasoning to persist, even when parts of the knowledge base are contradictory, by isolating and focusing on consistent fragments. In a Hybrid MKNF knowledge base, contradictions occur when both an atom $H$ and its classical negation $\neg H$ are derivable. For example, consider adding the two assertions \textit{OpnRwy(rw1)} and \textit{cldRwy(rw1)}. According to DL axiom (5), these assertions are contradictory, as a runway cannot be both operational and closed simultaneously.  
After applying knowledge base doubling and the semi-negative transformation (Step 1 to Step 6 in figure 1), these assertions are translated as follows:

\vspace{-2.8em}
\begin{multicols}{2}
\begin{align}
\text{OpnRwy}(rw1).  \\
\text{CldRwy}(rw1).
\end{align}

\columnbreak

\begin{align}
\text{OpnRwy}^d(rw1) \leftarrow \mathbf{not}  \  \text{NOpnRwy}(rw1).  \\
\text{CldRwy}^d(rw1) \leftarrow \mathbf{not}  \  \text{NCldRwy}(rw1).
\end{align}
\end{multicols}
\vspace{-.2em}
Using rule (11), we can infer \textit{Rwy(rw1)}. However, according to rule (12), we cannot 
infer $\text{Rwy}^d(rw1)$ because \textit{NOpnRwy(rw1)} holds, as established by rule (14).  A literal 
$H$ is considered \emph{consistently true} only if both $H$ and $H^d$ are true. In this case, although
$\text{Rwy}(rw1)$ is true, $\text{Rwy}^d(rw1)$ is false. Thus, the situation is treated as an 
inconsistency. Consequently, $\text{Rwy}(rw1)$ is excluded from the set of consistent query answers. By 
employing this syntactic approach, the NoHr reasoner effectively detects contradictions, ensuring that 
only consistent information is conveyed in response to queries, wherein an atom and its duplicate are 
considered true.

Below, we outline the consistency-resolving techniques employed by the NoHr reasoner. Consider a conjunctive query of the form:
\vspace{-.2em}
\begin{gather}q(X) \gets  A_1,   A_2, \dots,  A_n, \mathbf{not}  \ B_1, \mathbf{not}  \ B_2, \dots, \mathbf{not}  \ B_m\end{gather} 

where \(A_1, A_2, \dots, A_n, \mathbf{not}  \ B_1, \mathbf{not} \ B_2, \dots, \mathbf{not}  \ B_m\) represent the conjunction
of predicates being queried within the context of a Hybrid MKNF framework. To address the 
paraconsistent feature, the query is transformed into two distinct queries: 
\vspace{-.2em}
\begin{gather}
q(X) \gets  A_1, A_2, \dots,  A_n, \mathbf{not}  \ B^d_1, \mathbf{not}  \ B^d_2, \dots,\mathbf{not}\  B^d_m \\
q^d(X) \gets  A^d_1,  A^d_2, \dots, A^d_n, \mathbf{not} \ B_1, \mathbf{not} \ B_2, \dots, \mathbf{not} \ B_m  \ 
\end{gather}

Consistent answers were identified by querying the Prolog engine with:
\(
q(X) \land q^d(X)
\) while
contradictory or inconsistent answers can be determined using:
\(
 q(X) \land \mathbf{not} \  q^d(X)
\).
This approach enables the system to manage contradictions by clearly separating consistent conclusions from contradictory ones, ensuring that the presence of contradictions does not undermine the soundness of the reasoning process.

\paragraph{Evaluation:} A comprehensive evaluation of the NoHr reasoner is essential for its integration into aeronautics applications, to assess its performance in real-world operational environments. This evaluation should include the identification of DL constructors supported by NoHr, which is particularly beneficial for knowledge engineers involved in ontology modeling. Additionally, it is essential to measure the time required for each preprocessing phase of the DL ontology and logic programs.  This determines the time required for the reasoner to become operational after new knowledge is introduced into either component or during a system reboot. Furthermore, examining the latency of query responses by the NoHr reasoner is crucial, as it must deliver results in less than a second due to the safety-critical nature of aeronautics, where delays in query answering can pose significant risks. Another important parameter is resource utilization; as previously mentioned, the reasoner may operate on constrained devices with limited resources. Therefore, the reasoner should not consume excessive memory or CPU resources.

To analyze the preprocessing delay of the DL component, we used the OWL2Bench benchmark, which is an extension of the LUBM dataset designed to generate DL ontologies of varying sizes across four OWL profiles: EL, QL, RL, and DL \cite{Singh2020OWL2BenchAB}. We used this benchmark to evaluate the DL part with the NoHr reasoner. For the logic program component, we employed the benchmark provided by the NoHr reasoner, which generates a number 'n' of rules based on an associated DL ontology. For each DL ontology generated from the OWL2Benchmark, utilizing the rule benchmark provided by the NoHr reasoner, rules are generated in quantities of 10, 100, 1,000, 10,000, and 25,000.   The rule benchmark is available here: \url{https://github.com/NoHRReasoner/NoHR/tree/master/nohr-benchmark} and OWL2Bench is available here: \url{https://github.com/kracr/owl2bench}. For query evaluation, we used a set of queries provided with the LUBM benchmark, making minor modifications to match the structure and vocabulary of the DL ontologies generated by the OWL2Bench benchmark. These were then converted into equivalent queries that were compatible with the Prolog engine. Queries are available here for reference: \url{https://swat.cse.lehigh.edu/projects/lubm/queries-sparql.txt}. To evaluate the memory consumption and CPU utilization, we used ontology files containing varying numbers of axioms.
\vspace{-2em}
\begin{figure}[htp]
    \begin{minipage}{0.32\textwidth}
        \includegraphics[width=1\linewidth]{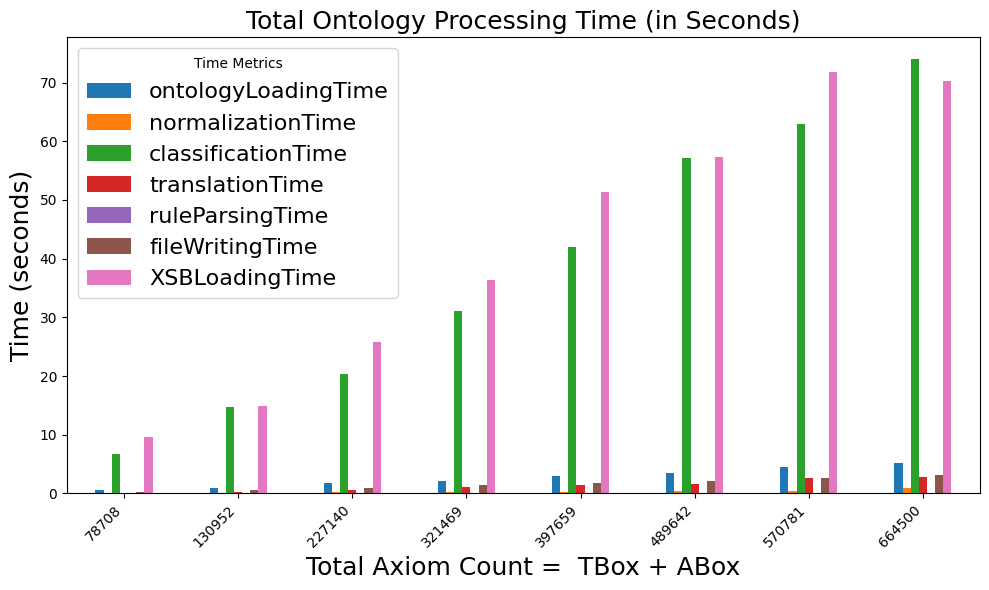}
        \caption{ Ontology Processing}
    \end{minipage}
     \begin{minipage}{0.32\textwidth}
      \includegraphics[width=\linewidth]{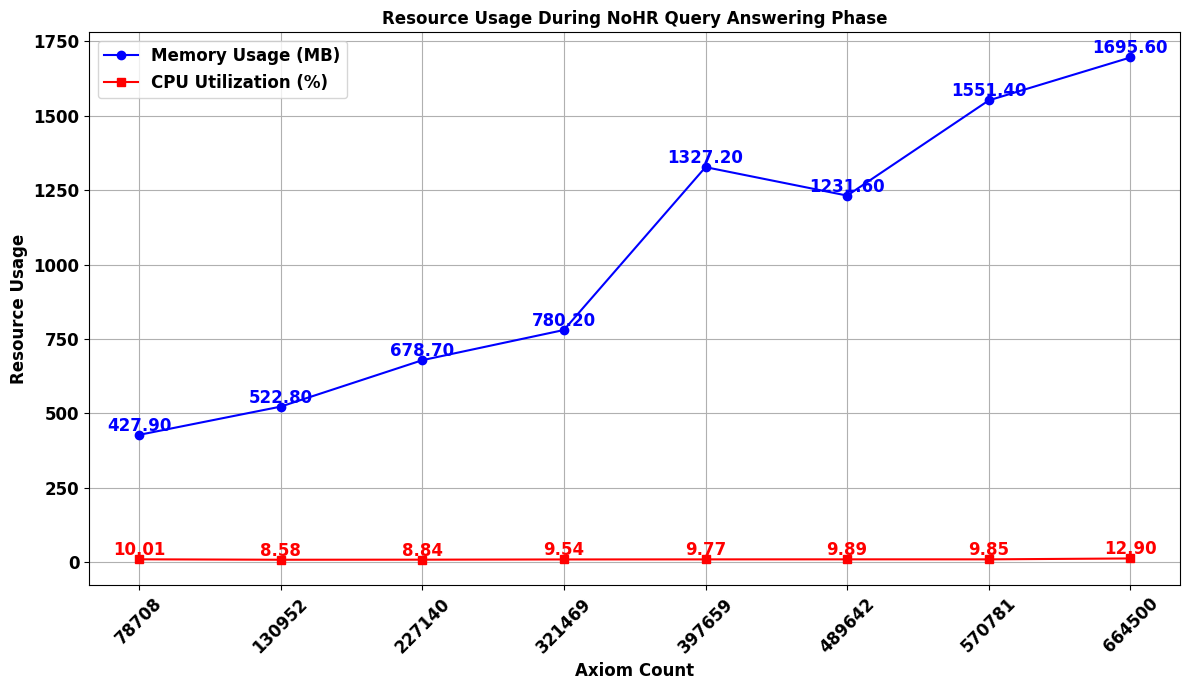}
       \caption{  Resource consumption}
    ~~\end{minipage}
     ~~\begin{minipage}{0.32\textwidth}
        \includegraphics[width=\linewidth]{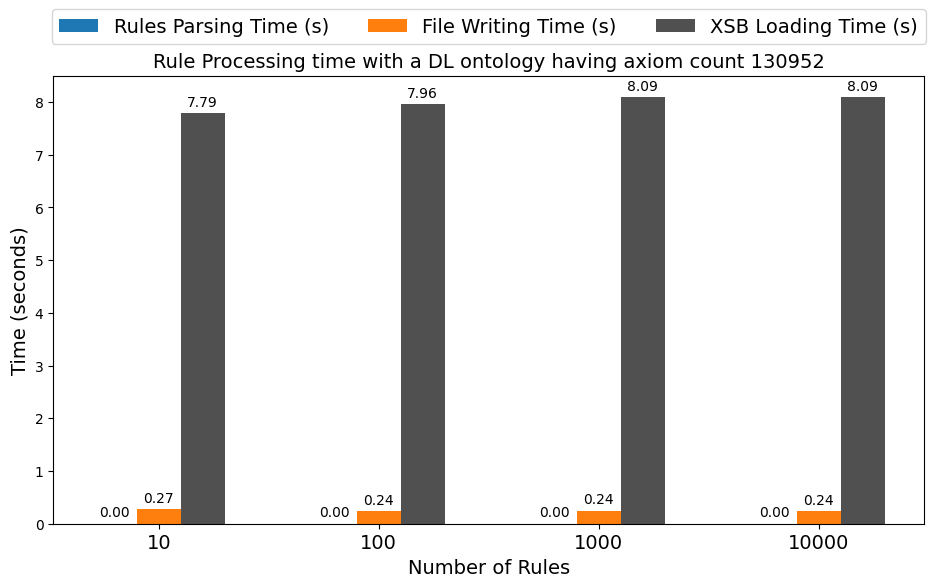}
      \caption{  Rule Processing (78708 DL axioms)}
    \end{minipage}
      \begin{minipage}{0.48\textwidth}
       
         \includegraphics[width=\linewidth]{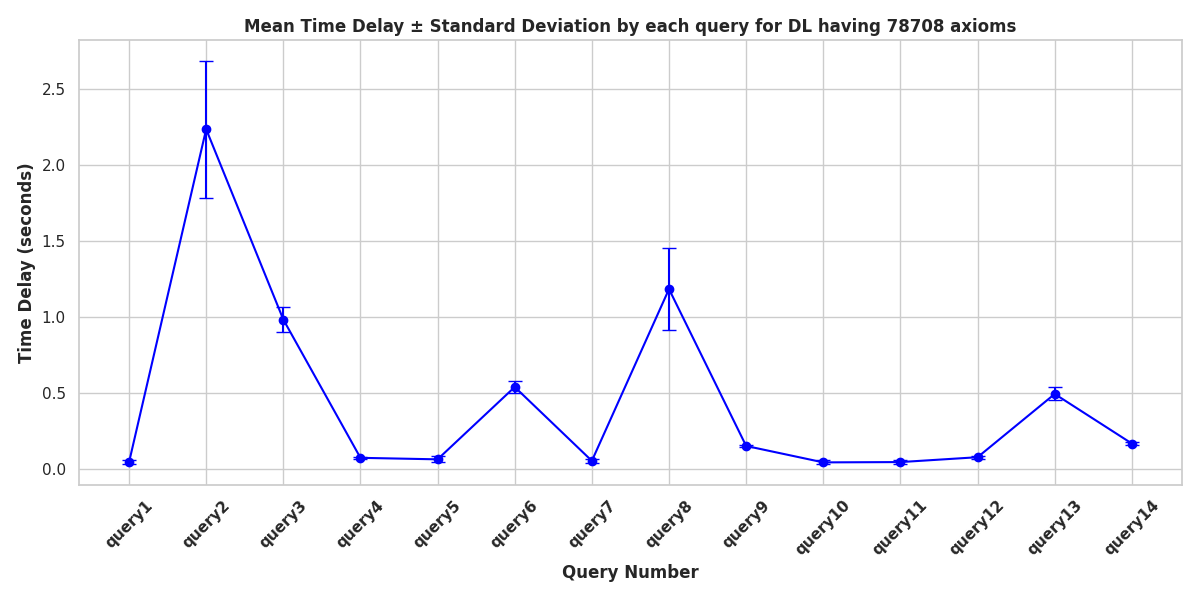}
              \caption{  Average Query response time for \\ 14 queries with a DL having 78708 axioms}
    \end{minipage}
      \begin{minipage}{0.48\textwidth}
       
         \includegraphics[width=\linewidth]{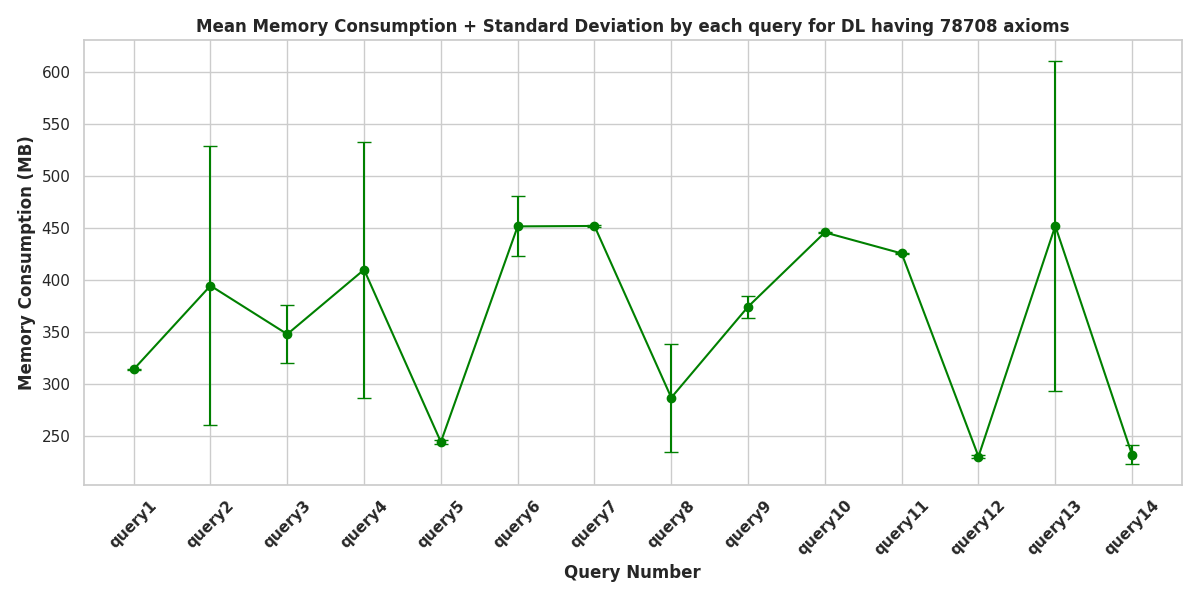}
     \caption{  Average Memory Consumption for each of the 14 queries on a DL having 78,708 axioms.}
    \end{minipage}
  \caption{Evaluation Result of NoHr Reasoner}
  \vspace{-1.2em}
\end{figure}

\paragraph{Evaluation Summary:}During the ontology preprocessing phase, a considerable amount of time is devoted to ontology classification and XSB loading, both of which are significantly affected by the number of DL axioms (figure 2). In the rule processing stage, the XSB loading time is directly proportional to the size of the translated ontology (figure 4). During preprocessing, we also noted increased memory usage and CPU utilization, both of which were correlated with the size of the knowledge base (figure 3). Once preprocessing is complete, query answering with XSB-Prolog becomes highly efficient, particularly for selective queries that filter results based on specific conditions and return only a small subset of answers (figure 5). However, we observed a noticeable decline in the performance of queries that required reasoning over recursive rules. Certain DL constructors, such as \textit{Equivalence Relationships} and \textit{Transitive Properties}, are translated into recursive rules, which significantly affect performance. Furthermore, elevated memory usage persisted in the query-answering phase (figure 6).
\vspace{-.3em}
\subsection{Usage}
Excessive memory and CPU consumption by reasoners can pose significant challenges in aeronautics applications, where real-time performance and resource efficiency are critical. In this context, the use of the NoHr reasoner with larger knowledge bases becomes problematic due to the high memory demands observed during evaluation. NoHr operates by translating a Hybrid MKNF knowledge base into an equivalent logic program using a sequence of transformation algorithms. This logic program is then executed by the XSB-Prolog engine to perform query answering. To mitigate memory consumption, we propose using NoHr exclusively during a preprocessing phase to generate the transformed logic program. During system operation, only the resulting logic program is executed independently with XSB-Prolog. This approach significantly reduces runtime memory usage while maintaining reasoning capabilities, as the live system no longer depends on the full NoHr infrastructure. As shown in figure 1, the steps up to the file writing phase are performed externally to generate the translated file. Subsequently, this file is used directly within a Prolog environment during the live application to execute the use case.

An essential component in implementing the proposed strategy is managing the consistency-resolving mechanism, which utilizes knowledge base doubling and semi-negative transformations. Although this process is internally handled by the NoHr reasoner, which is capable of returning both consistent and contradictory results, our approach relies exclusively on the translated, doubled logic program executed independently within a Prolog engine. Therefore, a custom strategy must be developed within the Prolog environment to replicate the behavior originally facilitated by NoHr.

\section{Additional Expressive Features}
By analyzing various use cases, we identified the need for additional expressive features
crucial for implementing aeronautics applications that are not supported in Hybrid MKNF by default. One such feature is
the explicit representation of negative knowledge within the LP component
of Hybrid MKNF. Consider the following case: "\textit{A runway is operational only if 
there is no NOTAM indicating that it is closed, and no obstacle is present on the runway}".
This can be initially represented by the following MKNF rule:
\vspace{-.3em}
\begin{equation} 
\mathbf{K} \ \text{OpnRwy(X)} \gets \mathbf{K}\ \text{Rwy(X)},\ \mathbf{not}\ \text{ob(A,X)},\ \mathbf{not}\ \text{CldRwy(X)}. 
\end{equation}
However, due to the safety-critical nature of aeronautics, the predicate \textit{ob(A, X)}, which indicates that there exists an obstacle on runway \(X\) in airport \(A\), carries particular importance. Using \textbf{not} 
operator with \textit{ob(A,X)} to infer that a runway is operational introduces safety concerns because it 
relies on the absence of evidence rather than explicit confirmation. Such assumptions are 
unacceptable in this domain. Instead, we must explicitly verify that the runway is free of
obstacles, thereby enhancing the reliability of the system. One way to achieve this is by 
replacing the \textbf{not} operator with classical negation denoted as \(\neg\). 

Classical negation can be incorporated into existing Hybrid MKNF semantics via syntactic transformation for unary and binary predicates. For a unary predicate \( A \), we introduce a new concept \( B \) with the axiom \( B \equiv \neg A \) in DL, allowing \( B \) to stand for the classical negation of \( A \) within rules. For a binary predicate \( P \), a corresponding predicate \( Q \) is created, and the disjointness axiom 
\(
P \sqcap Q \sqsubseteq \bot
\)
enforces that \( Q \) represents the classical negation of \( P \). However, this approach requires additional knowledge engineering 
and is limited to unary and binary predicates. So extending the rule part with a classical negation operator enhances expressiveness and also simplifies knowledge engineering and also allows for easier modeling of critical 
scenarios similar to the example provided above. After the extension of classical negation, the rule (22) can be 
rewritten as:
\vspace{-.3em}
\begin{equation}
    \mathbf{K} \ \text{OpnRwy(X)} \gets \mathbf{K} \ \text{Rwy(X)},\mathbf{K} \neg\text{ob(A,X)},\ \textbf{not}\ \text{CldRwy(X)}.
\end{equation}
Another crucial feature currently lacking is the capability to represent integrity constraints within the knowledge base and enforce them during query answering.  In the following sections, we present heuristics for implementing classical negation and discuss a method to ensure that the query results comply with all specified integrity constraints within the context of Hybrid MKNF.
\vspace{-.3em}
\subsection{Classical Negation}
Building on the insights from \cite{classicalnegationsyntactic} and \cite{classicalnegationarticle}, along with existing contradictory resolution techniques, such as knowledge base doubling and semi-negative transformation, we propose a method to integrate classical negation into the LP component of a Hybrid MKNF knowledge base. In this context, the modified Hybrid MKNF is represented as \(KB = (O, P)\), where \(O\) denotes the DL ontology and \(P\) comprises the rules of form (1). Here, \(H\), \(A_m\) for \(1 \leq i \leq m\) and \(B_n\) for \(1 \leq j \leq n\) are first-order atoms that may be preceded by classical negation, as indicated by (\(\neg\)). In this revised version of the  Hybrid MKNF, the LP rules now incorporate classically negative literals. As highlighted in \cite{classicalnegationsyntactic}, this classical negation can be eliminated through syntactic transformation, enabling the reuse of existing Hybrid MKNF algorithms for reasoning, which is the core concept of this proposal. The modified knowledge base \(KB^\prime\) is obtained from \(KB\) by substituting each classically negated atom \(\neg L\) in P with a new predicate \(L^\prime\) and introducing constraints \(C\) in the form \(\bot \gets \mathbf{K} \ L , \mathbf{K} \ L^\prime\). This results in a modified knowledge base, \(KB^\prime = (O, P^\prime \cup C)\). Subsequently, knowledge base doubling and semi-negative transformation are applied, as detailed in \cite{Alferes2010QueryDrivenPF}, leading to the doubled knowledge base \(KB^{\prime^d} = (O, O^d, P^{\prime^d})\), as outlined below:
\begin{enumerate}
    \item \( O^d \) is obtained by substituting each predicate \( L \) in \( O \) with \( L^d \);
    \item \( P^d \) is derived by transforming each rule of the form (1) occurring in \( P \) into two rules:
    \begin{enumerate}
        \item[i] \( \mathbf{K} \ H \leftarrow \mathbf{K} \  A_1, \dots, \mathbf{K} \ A_i, \textbf{not} \ B^{d}_1, \dots, \textbf{not} \ B^{d}_j \);
        \item[ii] \( \mathbf{K} \ H^d \leftarrow \mathbf{K} \  A^{d}_1, \dots, \mathbf{K} \  A^{d}_i, \textbf{not} \ B_1, \dots, \textbf{not} \ B_j, \textbf{not} \ N H \);
      \end{enumerate}

      \item \(C^d\) is derived from each constraint of the form \(\bot \gets \mathbf{K} \ L,\textbf{K} \ L^\prime\) in C is transformed into two rules: \(\mathbf{K} \ NL \gets \mathbf{K} \  L^\prime\) and \(\mathbf{K} \ NL^\prime \gets \mathbf{K}\ L\).
        (\(\bot \gets\mathbf{K} \ L, \mathbf{K} \ L^\prime \equiv \mathbf{K} \ \neg L \gets \textbf{K} \ L^\prime\) and \(\mathbf{K}\ \neg L^\prime \gets \textbf{K} \ L\). Replace \(\neg L\) and \(\neg L ^\prime\) with \(NL\) and \(N L ^\prime\)  .)
\end{enumerate}
where  \(A_i\), \(A^d_i\), \(B_j\), \(B^d_j\), \(H\), and \(H^d\) may represent either \(L\) or \(L^\prime\). 

Consider Hybrid MKNF knowledge base \(KB_2= (O, P)\), where \(O\) contains an assertion \textit{Rwy(rw1)} and \(P\) includes rules 
(23) and a fact: \(\mathbf{K} \  \neg \textit{ob(ifbo,rw1)}\). According to the syntactic transformation, these rules can be converted into the following rules, resulting in the modified knowledge base \(KB_2^\prime = (O, P^\prime \cup C)\) as illustrated below (only the LP part is provided):
\vspace{-.3em}
\begin{align}
\mathbf{K} \  \text{OpnRwy(X)} \gets \mathbf{K}\ \text{Rwy}(X),\mathbf{K} \ \text{ob}^\prime(A,X),\ \textbf{not}\ \text{CldRwy}(X).
\end{align}
\vspace{-4.5em}
\begin{multicols}{2}
\begin{align}
 \bot \gets \mathbf{K} \ \text{ob}(A,X),  \mathbf{K} \ \text{ob}^\prime(A,X).
\end{align}

\columnbreak

\begin{align}
\mathbf{K} \ \text{ob}^\prime(lfbo,rw1). 
\end{align}
\end{multicols}

After knowledge base doubling and semi-negative transformation we get \(KB_2^{\prime^d} = (O, O^d, P^{\prime^d})\): 
\begin{gather}
\mathbf{K} \  \text{opnRwy(X)} \gets \mathbf{K}\ \text{rwy}(X),\mathbf{K} \ \text{ob}^\prime(A,X),\ \mathbf{not}\ \text{cldRwy}^d(X).\\
\mathbf{K} \   \text{opnRwy}^d(X) \gets \mathbf{K}\ \text{rwy}^d(X),\mathbf{K} \ \text{ob}^{\prime^d}(A,X),\ \mathbf{not}\ \text{cldRwy}(X), \mathbf{not} \ \text{NopnRwy}(X).
\end{gather}
\vspace{-4.5em}
\begin{multicols}{2}
\begin{align}
 \mathbf{K}  \ \text{Nob}(A,X) \gets \mathbf{K} \  \text{ob}^\prime(A,X). \\
 \mathbf{K} \  \text{Nob}^\prime(A,X) \gets \mathbf{K} \  \text{ob}(A,X). 
\end{align}

\columnbreak

\begin{align}
\mathbf{K} \ \text{ob}^\prime(lfbo,rw1).  \\
\mathbf{K} \ \text{ob}^{\prime^d}(lfbo,rw1)\gets \mathbf{not} \ \text{Nob}^\prime(lfbo,rw1) . 
\end{align}
\end{multicols}
After applying all transformations, we obtain the modified knowledge base \(KB_2^{\prime^d} = (O, O^d, P^{\prime^d})\), where all occurrences of classical negation within the rules are syntactically eliminated. Despite the introduction of additional predicates and rules, we hypothesize that the proposed syntactic transformation preserves the semantic equivalence between the original Hybrid MKNF knowledge base with classical negation and the transformed negation-free knowledge base. This hypothesis is supported by existing approaches, such as \cite{classicalnegationsyntactic} and \cite{classicalnegationarticle}, which follow the same transformation to address classical negation within a logic program, as well as the knowledge base doubling and semi-negative transformation proposed in \cite{Alferes2010QueryDrivenPF} to handle classical negation within the DL part of Hybrid MKNF. Formal proof of semantic equivalence is the subject of ongoing work and is planned for future publications. Consequently, we can employ existing query answering algorithms such as those proposed in \cite{Ivanov2013AQTEL} and \cite{Costa2015NextSFQL}. This is achieved by converting the DL part into its semantically equivalent LP rules \(P^O\), then applying syntactic transformation to remove classical negation, \(KB_2^\prime = (\emptyset,P^O, P^\prime \cup C)\), and subsequently apply knowledge base doubling and semi-negative transformation, and use the resulting program to perform query answering using XSB-Prolog.

By adding the assertion \textit{rwy(rw1)} in  \(KB_2\), we can infer \textit{opnRwy(rw1)} because there is an explicit negative fact that indicates the absence of an obstacle, specifically \(\neg \textit{ob(lfbo,rw1)}\), and no closure is indicated. Now, consider a scenario of contradiction, which is an additional assertion \textit{ob(lfbo,rw1)} that also needs to be doubled.  In this case, \textit{ob(lfbo,rw1)} becomes contradictory  because both the atom and its classical negation are true simultaneously. In such a situation, the rules defined in (29) and (30) ensure that both \(N\mathit{ob}(X,Y)\) and \(N\mathit{ob}^\prime(X,Y)\) are true. Consequently, because of the semi-negative transformation, both \(\mathit{ob}^{\prime^d}(lfbo, rw1)\) and \(\mathit{ob}^d(lfbo, rw2)\) are inferred to be false. This allows the detection of atom  \textit{ob(lfbo,rw1)}, along with its duplicate \(\textit{ob}\prime\textit{(lfbo,rw2)}\), to be contradictory. Recall that an atom \(L\) is considered consistently true only if both \(L\) and its duplicate \(L^d\) are true. Then, the consistent query answering mechanisms described earlier can be used to correctly handle such situations and provide only consistent answers.
\vspace{-.3em}
\subsection{Integrity Constraints}
Integrity constraints are conditions used to verify the validity of both explicit and inferred knowledge. They are not intended to produce new inferences, but rather to ensure data consistency \cite{integrityconstraintsofrmaldefintion}. Our objective is to explicitly represent these constraints and ensure that all atoms in the result of a given query comply with them.  Any atom found to violate a constraint is identified as a constraint violation and is excluded from the final query response. Inspired by the work in \cite{activeintegrityexplicit} and \cite{activeintegrityconstraints}, we propose representing integrity constraints as condition-action rules of the form:
\vspace{-.4em}
\begin{align}
   A_1,\dots, A_n,\ \text{not}\ B_1, \dots, \text{not}\ B_m \Rightarrow a_1,\dots,a_k
\end{align}
Here, \(A_1,\dots,A_n\) and \(\text{not}\ B_1, \dots, \text{not}\ B_m\) represent the condition, while \(a_1,\dots,a_k\) denote the actions. Each action is represented by an atom \( \textbf{K} \  A \) or its classical negation \( \textbf{K} \ \neg A \).
Consider a fact that is deemed consistent only if a specific condition is satisfied. When this condition is not satisfied, the atoms responsible for the violation are identified as conflicting with the defined constraints and are excluded from subsequent reasoning processes. This is achieved by representing the violated condition as a rule of the form (33), where the head contains the classical negation of the atoms to be marked as conflicting.  When the body of the rule holds, it signifies a constraint violation, and the corresponding rule head is derived. As a result, both the original atoms and their classical negation may coexist in the knowledge base, leading to inconsistencies.  The rule (33) is converted into a set of MKNF rules with \(\mathbf{K}\) and \( \mathbf{not}\) operator as below:
\vspace{-.5em}
\begin{align*}
   \mathbf{K} \ a_1 \gets  \mathbf{K} \ A_1,\dots,\mathbf{K} \  A_n,\ \textbf{not}\ B_1, \dots, \textbf{not}\ B_m  \\
   ... \\
    \mathbf{K} \ a_k \gets  \mathbf{K} \ A_1,\dots,\mathbf{K} \ A_n,\ \textbf{not}\ B_1, \dots, \textbf{not}\ B_m 
\end{align*}

After the conversion, the constraint is treated as a standard MKNF rule. To handle classical negation, we apply the proposed syntactic transformation techniques and incorporate an inconsistency resolution strategy based on knowledge base doubling and semi-negative transformation. When the body of the translated MKNF rule evaluates to true, the head infers the negation of an atom, indicating a violation. This leads to contradictions in the MKNF knowledge base, where both an atom and its negation are simultaneously present. Consequently, by applying the consistency resolution strategy only consistent answers are produced during query answering, thereby removing all contradictory answers. This feature has several  applications in the aeronautics domain. One such application involves identifying incomplete NOTAMs, such as those missing critical information like the start time, end time, or issuing agency. These incomplete messages are considered inconsistent and are excluded from the reasoning process to preserve the integrity of decisions made by the system. An example of an integrity constraint that enforces a basic validity condition on NOTAM messages is: "\textit{A NOTAM must include both a start time and an end time}".
\begin{equation}
    \mathbf{K} \  \neg \mathsf{Notam}(X) \leftarrow \mathbf{K} \ \mathsf{Notam}(X), \ \mathbf{not}\ \mathsf{hasStartTime}(X, Y), \ \mathbf{not}\ \mathsf{hasEndTime}(X, Z).
\end{equation}
According to rule (34), if either the start time or end time is absent, the NOTAM is deemed inconsistent and is consequently exclude them from the further reasoning process. Another application is the detection of contradictory NOTAMs, for example, when two NOTAMs issued by different agencies refer to the same runway during the same time period, but one indicates that the runway is closed while the other states that it is operational.  Integrity constraints with repair action will help to identify such NOTAMs and excluded from further reasoning process.

\section{System Description}
Figure 8 illustrates the simplified architecture of the system. We developed a preprocessing tool that utilizes the translator embedded in the NoHr reasoner to transform DL ontology into an equivalent set of rules. This process involves knowledge base doubling and semi-negative transformation. During translation, the NoHr translator systematically mapped each concept and role from the DL ontology into two distinct predicates representing both the original and doubled versions.
\begin{figure}[htp]
    \centering
    \includegraphics[width=.7\linewidth]{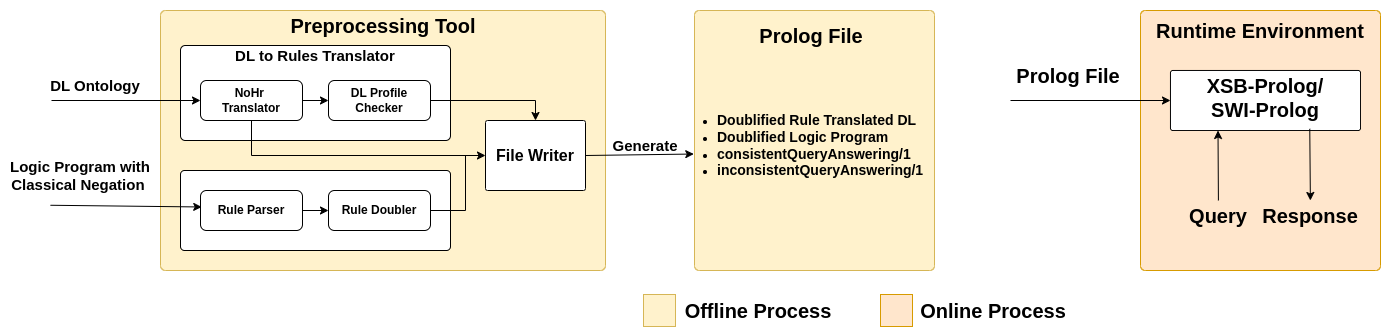}
    \caption{Architecture of Proposed Work Flow }
    \label{fig:enter-label}
    \vspace{-1.4em}
\end{figure}
For example, \textit{Rwy} is mapped to \textit{aRwy} to represent the original predicate, and to \textit{dRwy} to represent the doubled predicate. Furthermore, all predicates in the translated program are declared tabled, and the standard \textbf{not} operator is replaced with tabled negation, denoted by \textbf{tnot}. For example, DL axiom (4) is translated into rules (11) and (12), which appear in the Prolog program below, as generated by the NoHr translator.

\begin{Verbatim}[numbers=left]
arwy(X):- aopnRwy(X).
drwy(X):- dopnRwy(X), tnot(nrwy(X)).
\end{Verbatim}

Classical negation in rules can be represented using the symbol "\textit{-}". An additional parser is required to handle this classical negation in the LP part of Hybrid MKNF. This parser eliminates classical negation according to the proposed syntactic transformation technique, and subsequently applies both knowledge base doubling and semi-negative transformation.   Furthermore, it designates all atoms as tabled, and alters the \textbf{not} operator to \textbf{tnot}. The rule parser converts rules (27) and (28) into a prolog program as follows (see Section 6.1): $\neg\textit{ob(X)}$ is substituted with a newly introduced predicate $\textit{nonob(X)}$ by adding a prefix "\textit{non}" instead of \(\neg\).  

\begin{Verbatim}[numbers=left]
aopnRwy(X):- arwy(X),anonob(A,X),tnot(dcldRwy(X)).
dopnRwy(X):- drwy(X),dnonob(A,X),tnot(acldRwy(X)),tnot(nopnRwy(X)).
\end{Verbatim}

To implement query answering, two meta-predicates are introduced: \textit{consistentQuery/1}, which returns consistent answers, and \textit{inconsistentQuery/1}, which returns contradictory answers. These meta-predicates implement  query doubling, as explained in (20) and (21), by  
matching them with  transformed logic programs.  After doubling, these meta-predicates implement  the conditions \( q(X) \land q^d(X) \) and  \( q(X) \land \textbf{not }\, q^d(X) \) to query the consistent and contradictory answers, respectively. The built-in Prolog predicate 
\textit{atom\_concat/3} is employed to prepend additional strings, which is beneficial for query doubling. Subsequently, the \textit{call/1} predicate is used to execute the transformed queries. The \textit{file writer module} integrates translated DL with predefined meta-predicates 
and doubled logic program from parser, exports to a file executable  by Prolog engines. Also it  generates a report on unsupported axioms in the DL ontology, with examples available at \url{https://drive.uca.fr/d/040c6ecbe940449e9c0d/}. After translation, the doubled LP part along 
with the translated DL axioms can be executed using a Prolog engine that supports well-founded semantics. For query answering, we used SWI-Prolog or XSB-Prolog for evaluation. figures 9 and 10 illustrate the query response time and memory consumption during the evaluation of each 
query for XSB-Prolog and SWI-Prolog, respectively. Overall, we observe a substantial reduction in memory usage across all queries, except for the first execution of more complex recursive queries. This is because the tabling feature stores previous results after the initial computation, allowing subsequent executions of the same query to reuse these results without repeating the full computation. Preprocessed files and queries used for this evaluation are available here: \url{https://drive.uca.fr/d/e5482dd1af5f4fdd8bfb/}.
 
 \begin{figure}[htp]
     \begin{minipage}{0.49\textwidth}
        \includegraphics[width=\linewidth]{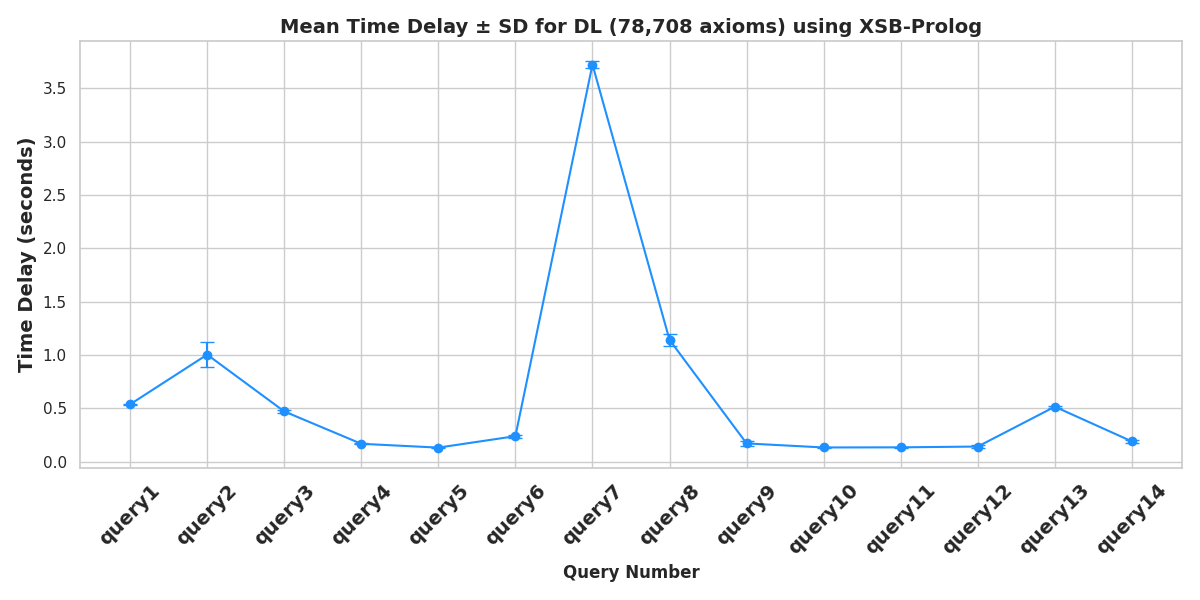}
    \end{minipage}
      \begin{minipage}{0.49\textwidth}
       \includegraphics[width=0.9\linewidth]{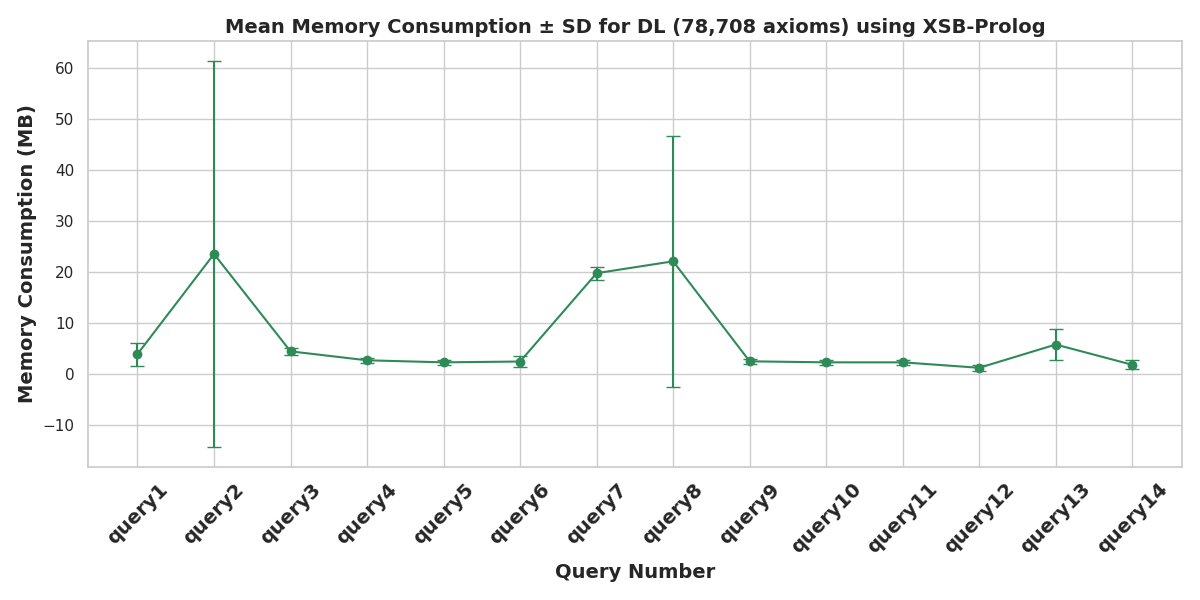}
        
    \end{minipage}
  \caption{ Response Time and Memory Consumption for each query with XSB-Prolog}
\end{figure}

\begin{figure}[htp]
     \begin{minipage}{0.49\textwidth}
        \includegraphics[width=\linewidth]{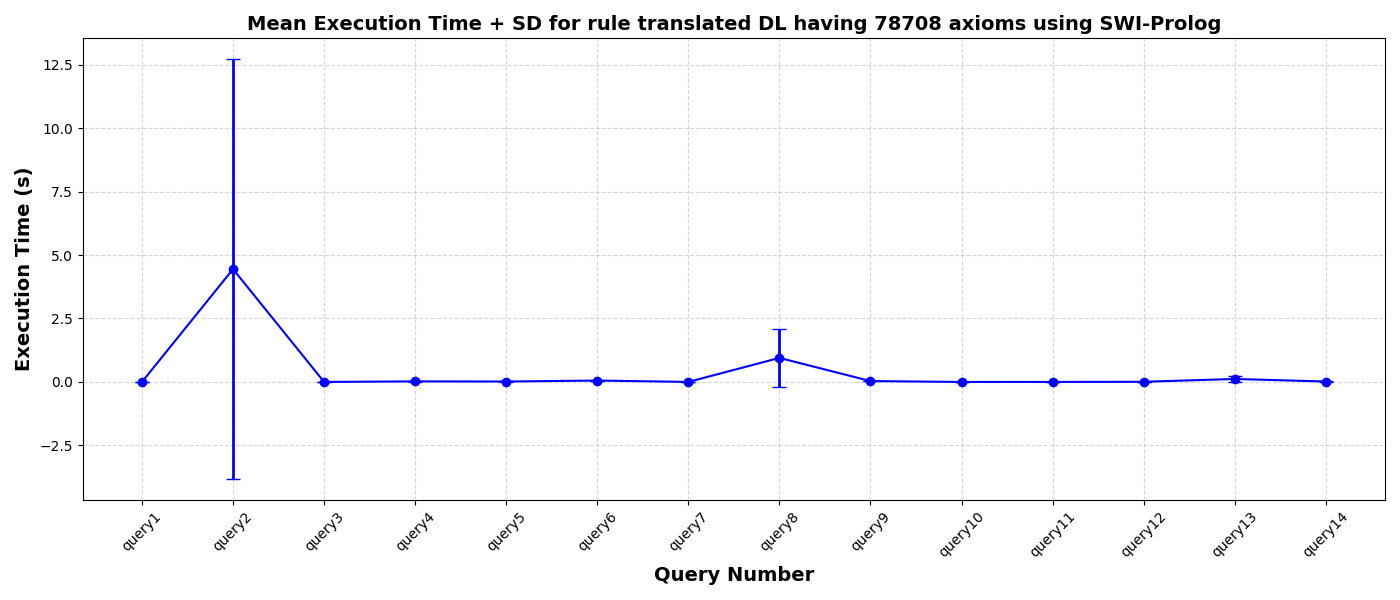}
    \end{minipage}
      \begin{minipage}{0.49\textwidth}
       \includegraphics[width=0.9\linewidth]{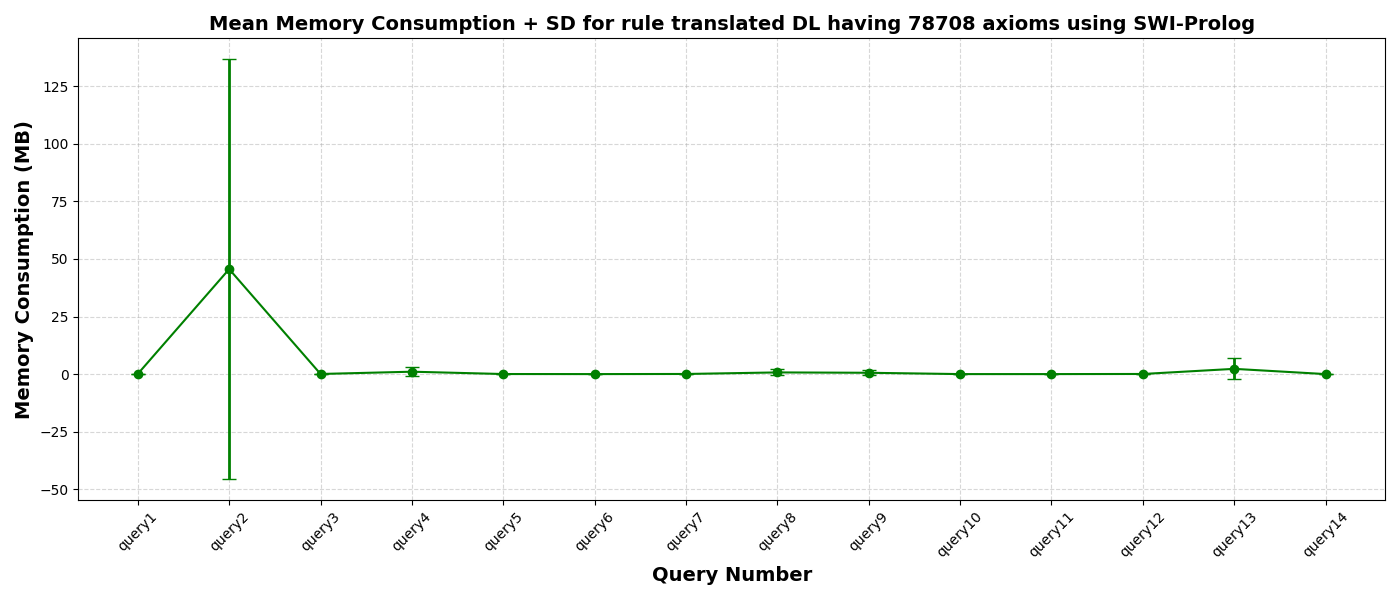}
        
    \end{minipage}
  \caption{Response Time and Memory Consumption for each query with SWI-Prolog}
  \vspace{-1em}
\end{figure}

\section{Related Work}

In general, three main approaches exist for combining DLs with LP: loose integration, tight integration, and full integration \cite{eiterrules2008}. Each of these integration strategies can be based on one of two major semantics: well-founded \cite{Gelder1991TheWS} or answer set semantics \cite{Eiter2009AnswerSP}. In loose integration, a DL ontology and LP rules are regarded as two separate entities, with a specified interface facilitating the exchange of knowledge. DL-Programs is a loose integration approach that proposed under both answer set semantics \cite{dl-program} and well-founded semantics \cite{Eiter2004WellfoundedSF}.  In DL-Programs, knowledge exchange between DL and LP rules is facilitated only through external operators, necessitating additional rules that increase the language's complexity for knowledge engineers. The only reasoning task in DL-Programs is model computation, where the resulting model is stored in a database to enable query answering.  However, this approach is not memory-efficient, as it requires computing and storing the full model. Nonetheless, DL-Programs offer the benefit of integrating with external data sources, such as Python programs or relational databases. An extended language known as \textit{HEX} \cite{Eiter2005AUI}, along with its corresponding reasoner, \textit{DLV-Hex} \cite{Eiter2005DLVHEXD}, supports this integration. There also exists an optimized reasoner called Drew System \cite{drew} that utilizes two datalog rewritable DLs,  \(\mathcal{LDL+}\) \cite{ldl+} and \(\mathcal{SROEL}\) \cite{sroel} for DL-Programs. Additionally, an alternative loose integration approach called F-Logic\# has been proposed that combines F-logic  with DL ontologies \cite{flogic}.

In tight integration, an integrated model that satisfies both the DL and LP parts is defined. 
The integrated model is \( M = M_O \cup M_L \), where \( M_O \) satisfies the DL ontology 
and \( M_L \) satisfies the corresponding LP rules.  We start with a model of the DL part, denoted as \(M_O\). Using this model, we simplify the grounded LP part in two steps. First, we remove any LP rules that contain DL atoms in their body that are false in \(M_O\), since those rules cannot be satisfied. Then, from the remaining rules, we remove any DL atoms in the body that are true in \(M_O\), as they are already satisfied. This results in a version of the LP program that no longer includes any DL atoms. We then compute a model for this simplified LP using either answer set  or well-founded semantics. This process is repeated for each possible DL model \(M_O\), producing a corresponding LP model for each one. Several hybrid languages have been developed with a focus on tight integration. Notably, R-Hybrid KB \cite{rhybridkb} and DL+Log \cite{Rosati2006DLlogTI} are founded on answer set semantics, while HD-Rule is based on well-founded semantics \cite{Drabent2007HDrulesAH}. Furthermore, an extension of DL+Log has been introduced, which incorporates closed-world predicates within a DL ontology, leading to the creation of clopen knowledge bases \cite{clopen}. Additionally, Resilient Logic Programs (RLP) have been proposed to merge both tight and loose integration, where loose integration shares only logical consequences between DL and LP, while tight integration enables  knowledge sharing between individual models \cite{Lukumbuzya2020ResilientLP}. In tight integration, even when employing well-founded semantics, numerous possible models may exist, complicating the reasoning process. Consequently, constructing a practical reasoner under these conditions is challenging. Additionally, the knowledge engineer must possess an understanding of individual models of  DLs to generate the corresponding rule program due to the model-based integration between DL and LP, further complicating the knowledge engineering process.

In full integration, a singular unifying non-monotonic formalism is defined to encapsulate both the semantics of DLs and LP. Several non-monotonic extensions of first-order logic have been proposed, including default logic, epistemic logic, defeasible logic, and circumscription. Ideally, selecting one of these extensions to integrate both DL and rules constitutes the fundamental concept of this approach. In full integration, we encounter works grounded in Minimal Knowledge and Negation-as-a-failure (MKNF), Open-Answer Set Programming (OASP) \cite{Heymans2007OpenAS}, and Defeasible Logics \cite{Antoniou2007DRPrologAS}. We intentionally excluded a significant array of research  that merges DLs with first-order rules, such as SWRL \cite{Lan2004SWRLA}, AL-Log \cite{al-log}, Carin \cite{carin}, due to their lack of closed-world reasoning capabilities.  Given the undecidability of OASP, several decidable variants have been proposed. These include the F-Hybrid Knowledge Base, which integrates \(\mathcal{SHOQ}\) DL with a finite set of Forest Logic Programs (FLP) \cite{f-hybrdi}. These programs allow only unary and binary predicates and adhere to the forest-model property. Additionally, the G-Hybrid knowledge base combines \(\mathcal{DLRO^{-\{\leq\}}}\) with a finite set of ASP rules that comply with the guardness restriction, wherein each rule contains a guarded predicate, and all variables within the rules must appear in that predicate \cite{Heymans2003GHybridKB}. In this context,  Hybrid MKNF offers a more general and expressive language that integrates DLs and LP \cite{hybridMknfintroduction}.

\section{Conclusion}
We introduced Hybrid MKNF as a knowledge representation language tailored for aeronautics applications, such as NARS. To address memory constraints in aeronautic systems, we propose an alternative workflow that employs the NoHr reasoner through offline preprocessing. Our approach manages both consistent and contradictory query responses by using Prolog meta-predicates. We developed heuristics to enhance Hybrid MKNF with classical negation and to support integrity constraints through repair actions. A tool was created to convert a Hybrid MKNF knowledge base into an equivalent set of LP rules using NoHr by integrating these heuristics. In future research, we plan to formally define the integration of classical negation in the LP into Hybrid MKNF semantics under well-founded semantics. This study aims to identify syntactic fragments of Hybrid MKNF, such as definite and stratified programs, that are compatible with lightweight rule engines like I-DLV, Trealla Prolog, and Tau-Prolog. The goal is to enable efficient reasoning on resource-constrained platforms and to guide the selection of suitable rule engines based on expressivity requirements.
We will enhance explainability by adapting provenance-based techniques to Hybrid MKNF under well-founded semantics to provide reasoning justifications.  Finally, we  will incorporate temporal reasoning into Hybrid MKNF for scenarios such as NOTAMs, where facts are valid over specific time intervals, by extending the framework with temporal operators.

\end{document}